# On the long-term learning ability of LSTM LMs

Wim Boes[1], Robbe Van Rompaey[1], Lyan Verwimp[2],
Joris Pelemans[2], Hugo Van hamme[1] and Patrick Wambacq[1] *†

1- KU Leuven - ESAT
Leuven - Belgium

2- Apple Inc.
Cupertino - USA

**Abstract**. We inspect the long-term learning ability of Long Short-Term Memory language models (LSTM LMs) by evaluating a contextual extension based on the Continuous Bag-of-Words (CBOW) model for both sentence- and discourse-level LSTM LMs and by analyzing its performance.

We evaluate on text and speech. Sentence-level models using the long-term contextual module perform comparably to vanilla discourse-level LSTM LMs. On the other hand, the extension does not provide gains for discourse-level models. These findings indicate that discourse-level LSTM LMs already rely on contextual information to perform long-term learning.

## 1 Introduction

LMs usually estimate the probability of a word sequence $\boldsymbol{w} = (w_1, ..., w_T)$ by multiplying the conditional probabilities of each word given its preceding words.

Recurrent neural networks are often used for this purpose. Particularly, the LSTM LM [1, 2] shown in Figure 1 has been the starting point of many successful projects, as demonstrated by, e.g., Merity et al. [3].

This model approximates the probability of a word sequence $\boldsymbol{w}$ by sequentially processing the words therein: each word $w_t$ is converted into an embedding vector and fed to one or more LSTM layers. Next, the softmax function is applied to obtain the conditional probability of the next word $p(w_{t+1}|w_1, ..., w_t)$.

LSTM LMs are not inherently limited in their ability of modeling word dependencies: due to the feedback in the LSTM layers (through the feedback of state vectors), these models can theoretically handle all long-distance relations.

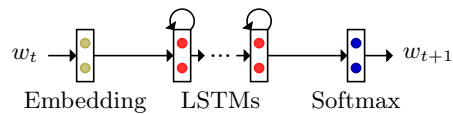

Fig. 1: LSTM LM

In practice however, the extent to which they can learn long-term dependencies seems to be restricted. Khandelwal et al. [4] suggest that LSTM LMs are able to remember only a limited amount of words preceding each input word.

---

*This work is supported by the Research Foundation Flanders (FWO-Vlaanderen).
†This work does not contain Apple Inc. proprietary information.

Also, Grave et al. [5] have shown that caches using a more remote word history strongly improve the performance of LSTM LMs, supporting this hypothesis.

In this work, we take a novel approach to investigating the long-term learning ability of LSTM LMs. We design a straightforward long-distance module, which uses the context of each input word to obtain relevant semantic information. This component is explicitly designed to be easily interpretable. We extend multiple LSTM LM configurations with this long-term contextual component and analyze performance disparities for both language modeling and speech recognition.

## 2 Long-term extension based on the CBOW model

The proposed long-term contextual extension to the LSTM LM is inspired by the CBOW system [6]. This model learns to predict each target word $w_t$ based on the words in its K-neighborhood $\{w_{t-K}, ..., w_{t-1}, w_{t+1}, ..., w_{t+K}\}$ by learning continuous representations of these words and log-linearly combining them. Mikolov et al. have shown that the learned embeddings show interesting linguistic properties. For instance, in the resulting vector space, singular-plural and male-female relationships are manifested through vectorial differences.

We try to exploit these useful properties to create a contextual feature vector $\boldsymbol{d_t}$ for predicting the word $w_{t+1}$ by transforming the words in its K-history $\{w_{t-K}, ..., w_t\}$ into embeddings $\{\boldsymbol{e_{t-K}}, ..., \boldsymbol{e_t}\}$ and combining them linearly:

$$\boldsymbol{d_t} = \frac{\sum_{k=0}^{K} g(k, w_{t-k}) \cdot \boldsymbol{e_{t-k}}}{\sum_{k=0}^{K} g(k, w_{t-k})} \qquad (1)$$

In (1), $g(k, w_{t-k})$ is the weight of $w_{t-k}$ in the combination. We propose three weighting schemes. Firstly, uniform weighting with $g(k, w_{t-k}) = 1$. Secondly, position-dependent or exponential weighting with $g(k, w_{t-k}) = \alpha^{-k}$. Here, $\alpha$ determines how much the weights of the words decay as we go further back in the history. Lastly, word-dependent weighting with $g(k, w_{t-k}) = IDF(w_{t-k})$, which is the inverse document frequency weight of $w_{t-k}$. In (1), this results in each unique word being weighted by its term frequency-inverse document frequency weight, a statistic reflecting the relative importance of the word [7].

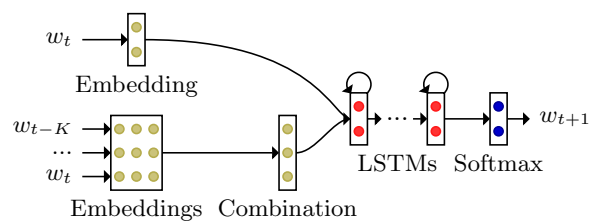

Fig. 2: LSTM LM with CBOW-based long-term extension

The feature vector $\boldsymbol{d_t}$ is inserted into the LSTM LM by concatenating it with the embedding of the current word $w_t$ before the LSTMs as shown in Figure 2.

## 3 Experiments

### 3.1 Setup

We use TensorFlow [8] to train LSTM LMs with and without extensions on the English WikiText-2 [9] and the Mediargus corpus, a Dutch in-house data set consisting of newspaper fragments from the Flemish digital press database Mediargus. The former corpus contains 2M training tokens and uses a vocabulary of 33K words, with an out-of-vocabulary (OOV) rate of 2.6%. The latter contains 130M training tokens and employs a vocabulary of 50K words (4% OOV).

Preliminary experiments are performed to tune the hyperparameters of the LSTM LMs. The resulting configurations are explained in the paragraphs below.

All LMs employ a single LSTM layer. The vanilla models use one embedding layer, while the ones with extension use two of which the weights are not shared.

For Wikitext-2, we train small/large LSTM LMs with 100/280-dimensional embeddings and 200/560 hidden neurons in the LSTM layer. In this case, the models with contextual modules contain more parameters than the vanilla models. For the Mediargus corpus, we train small/large vanilla LSTM LMs with 64/256-dimensional embeddings and hidden sizes of 128/512. We also optimize small/large models with long-term modules using 55/128-dimensional embeddings and hidden sizes of 100/512. In this case, the LSTM LMs with extensions contain roughly the same amount of neural weights as the vanilla models.

All neural weights are uniformly and randomly initialized in the range of [-0.05, 0.05], except for the word embeddings: they are either initialized as just outlined or with pretrained embeddings obtained via Word2Vec [6].

The CBOW-based extension explained in Section 2 utilizes the 100 most recent words to estimate the contextual feature vector, i.e., K is equal to 100 in (1). For position-dependent weighting, we choose $\alpha$ equal to 1.05. The IDF weights used for word-dependent weighting are extracted via gensim [10].

We use stochastic gradient descent in conjunction with backpropagation through time [11] to optimize the models. We apply gradient clipping [12] with a maximum gradient L2 norm of 5. During training, we apply dropout [13, 14] with a probability of 0.5 after the embedding and LSTM layers. The models use an initial learning rate of 1. For the English Wikitext-2, we use 75 epochs, a batch size of 20, 35 time steps for unrolling and apply exponential learning rate decay with a rate of 0.8 starting from epoch 6. For the Dutch Mediargus corpus, we use 3 epochs, a batch size of 50 and 50 time steps for unrolling.

The models are trained via either discourse-level training, where the hidden states of the LSTMs are propagated over sentence boundaries, or sentence-level training, where the hidden states are reset to zero after each sentence.

### 3.2 Qualitative experiments

We investigate whether the CBOW-based long-term module described in Section 2 is meaningful by feeding short text pieces with clear topics to trained LSTM LMs using the aforementioned extension and by analyzing their outputs.

We search for the words of which the embedding vectors $e$ are closest to the computed feature vector $d$ at the end of the text piece, expecting these words to contain relevant long-term information, i.e., to be representative of the context.

A small sample of the results is shown in Table 1 for the large discourse-level LMs with extensions trained on WikiText-2, using text snippets on very different topics, namely Tropical Storm Josephine, planet Upsilon Andromedae b and the military history of Australia, taken from the validation set of WikiText-2.

| Weighting | Snippet 1 | Snippet 2 | Snippet 3 |
|---|---|---|---|
| Uniform | depression | molecules | World |
|  | Tropical | helium | England |
| Positional (exponential) | storm | planet | States |
|  | tropical | core | United |
| Word-dependent | Tropical | star | conflicts |
|  | depression | magnetic | Australian |

Table 1: Qualitative evaluation

The words of which the embeddings are closest to the feature vectors are representative of the context, even though many do not appear in the used snippets, showing that the extension is capable of capturing long-term information.

### 3.3 Quantitative experiments

#### 3.3.1 WikiText-2

For the LSTM LMs trained on WikiText-2, we evaluate the extension by measuring the perplexity (PPL) the models produce on the test set of WikiText-2, which contains 250K tokens. Scores are averaged over 3 training runs to ensure the measurement variance is sufficiently small. The results are shown in Table 2.

Table 2 shows that for discourse-level LSTM LMs, the extension provides only marginal improvements, suggesting that the contextual information captured by the CBOW-based module is already present in these models. However, the test PPLs of sentence-level LSTM LMs are notably reduced: without contextual module, there is more than a 20% PPL difference between discourse- and sentence-level models. With the extension, this discrepancy drops to 7% or less.

| Extension | Weighting | PPL (discourse-level model/sentence-level model) | | | |
|---|---|---|---|---|---|
|  |  | Random emb. init. | | Pretrained emb. init. | |
|  |  | Small | Large | Small | Large |
| No | N/A | 117.5/142.4 | 97.6/118.0 | 111.1/134.3 | 93.9/114.6 |
| Yes | Uniform | 115.8/125.8 | 97.4/105.1 | 108.7/117.7 | 93.1/99.8 |
|  | Positional | 114.4/128.6 | 97.1/109.5 | 107.8/121.4 | 93.1/104.9 |
|  | Word-dep. | 118.7/127.2 | 97.7/107.5 | 109.3/118.9 | 92.8/100.7 |

Table 2: Quantitative evaluation on WikiText-2

*3.3.2 Mediargus corpus and Corpus Gesproken Nederlands (CGN)*

For models trained on the Mediargus corpus, we measure PPLs on the test set thereof (50K tokens). Also, we test the LMs in a speech recognition application. We use the models to perform 100-best list rescoring on component k of CGN [15], a Dutch speech data set containing news bulletins (41K tokens): we first use the ESAT 2008 speech recognizer [16] to return 100 possible hypotheses for each word list. These suggestions are then rescored by LSTM LMs with and without long-term contextual extensions, and the resulting word error rates (WERs) are measured. PPL and WER scores are averaged over 3 training runs to ensure the measurement variance is adequately reduced. The results are shown in Table 3.

| Model | Weighting | PPL/WER (%) | | | |
|---|---|---|---|---|---|
| | | Random emb. init. | | Pretrained emb. init. | |
| | | Small | Large | Small | Large |
| Discourse-level vanilla model | N/A | 238.1/19.31 | 135.1/18.74 | 217.0/19.30 | 122.3/18.62 |
| Sentence-level model with extension | Uniform | 222.5/19.31 | 132.3/18.76 | 213.3/19.26 | 122.2/18.61 |
| | Positional | 223.0/19.39 | 133.1/18.78 | 215.8/19.27 | 126.5/18.72 |
| | Word-dep. | 219.9/19.34 | 129.0/18.81 | 214.3/19.36 | 138.9/18.65 |

Table 3: Quantitative evaluation on the Mediargus corpus and CGN

Table 3 shows a similar trend to Table 2: in terms of both PPL and WER, sentence-level models with CBOW-based module perform similarly to discourse-level LMs without extension. For a more detailed analysis we refer to [17].

*3.3.3 Discussion*

Khandelwal et al. [4] have shown that (discourse-level) LSTM LMs grasp long-term information via vague semantic concepts of a restricted word history.

The CBOW-based extension, which is essentially a semantic representation of the context, does not improve discourse-level LSTM LMs. This result supports the finding of Khandelwal et al. that discourse-level LSTM LMs are inherently capable of learning long-term information up to a limited extent.

In addition, sentence-level LMs upgraded with the long-distance extension perform similarly to plain discourse-level LSTM LMs, indicating that the long-term semantic concepts learned by discourse-level LSTM LMs are even somewhat comparable to the considered CBOW-based contextual representations.

## 4 Conclusion

We inspected the long-term learning ability of LSTM LMs by investigating a long-term extension based on the CBOW model. We performed language modeling and speech recognition experiments on English and Dutch data sets.

The CBOW-based module lead to strong PPL reductions under sentence-level training but not under discourse-level training. The sentence-level LMs with extensions even performed comparably to vanilla discourse-level models.

The considered CBOW-based module apparently does not contain additional useful long-term cues for discourse-level LSTM LMs because the contextual information it carries is already implicitly present in their recurrent hidden states.

The results in this work indicate that LSTM LMs are inherently capable of learning basic semantic information of a limited history, i.e., the context. For future research, easy wins might therefore be realized with modifications which are not just semantic in nature. Additionally, it could be useful to investigate extensions that can handle the more remote history. This is also evidenced by the success of cache-based extensions: indeed, these extensions are not semantic but memory-based, and they utilize very distant history word histories.